\documentclass[conference]{IEEEtran}
\IEEEoverridecommandlockouts
\usepackage{cite}
\usepackage{amsmath,amssymb,amsfonts}
\usepackage{algorithmic}
\usepackage{graphicx}
\usepackage{textcomp}
\usepackage{xcolor}

\usepackage{subcaption}
\usepackage{multirow}
\usepackage{booktabs}

\def\BibTeX{{\rm B\kern-.05em{\sc i\kern-.025em b}\kern-.08em
    T\kern-.1667em\lower.7ex\hbox{E}\kern-.125emX}}

\newcommand{\x}{\mathbf{x}}
\newtheorem{definition}{Definition}
    
\begin{document}

\title{Knowledge-based Refinement of Scientific Publication Knowledge Graphs
\thanks{SY and SN gratefully acknowledge the support of AFOSR Minerva award FA9550-19-1-039. Any opinions, findings, and conclusion or recommendations expressed in this material are those of the authors and do not necessarily reflect the view of the AFOSR, or the US government.}
}

\author{\IEEEauthorblockN{Siwen Yan}
\IEEEauthorblockA{\textit{Computer Science Department} \\
\textit{University of Texas at Dallas}\\
Dallas, USA \\
Siwen.Yan@utdallas.edu}
\and
\IEEEauthorblockN{Phillip Odom}
\IEEEauthorblockA{\textit{Georgia Tech Research Institute} \\
\textit{Georgia Institute of Technology}\\
Atlanta, USA \\
phillip.odom@gtri.gatech.edu}
\and
\IEEEauthorblockN{Sriraam Natarajan}
\IEEEauthorblockA{\textit{Computer Science Department} \\
\textit{University of Texas at Dallas}\\
Dallas, USA \\
Sriraam.Natarajan@utdallas.edu}
}

\maketitle

\begin{abstract}
We consider the problem of identifying authorship by posing it as a knowledge graph construction and refinement. To this effect, we model this problem as learning a probabilistic logic model in the presence of human guidance (knowledge-based learning). Specifically, we learn relational regression trees using functional gradient boosting that outputs explainable rules. To incorporate human knowledge, advice in the form of first-order clauses is injected to refine the trees. We demonstrate the usefulness of human knowledge both quantitatively and qualitatively in seven authorship domains.
\end{abstract}

\begin{IEEEkeywords}
knowledge graphs, statistical relational learning, knowledge-based learning, authorship prediction
\end{IEEEkeywords}

\section{Introduction}
Building knowledge graphs (KG) that can then be allowed for reasoning has been a long dream of AI. A classic example is the Cyc system by Cycorp~\cite{Lenat95} which has accumulated knowledge for decades. KG construction has been addressed by different techniques including curation~\cite{Lenat95}, crowd sourced~\cite{Freebase,Wikidata}, or extracted from web~\cite{NELL,PROSPERA,KnowledgeVault}. As noted by Paulheim~\cite{PaulhierSurvey}, irrespective of the manner of the construction of KGs, the resulting KG is almost always never perfect. This necessitates the refinement of KGs. 

The difference between construction and refinement is that the latter assumes that a KG is present and would be corrected based on some (newer) data. While in principle, KG construction methods could be applied for refinement as well and vice-versa, the goals are different. Most importantly, KG refinement can be more generalized than construction and applied to different graphs, and understanding refinement methods could lead to better clarity on the effects of the operators~\cite{PaulhierSurvey}. In this paper, we focus on KG refinement problem.

We note that KGs are inherently relational (i.e., it contains objects, attributes of these objects and relationships between different objects) and noisy (the graph changes over time and newer relationships are added/deleted etc). Consequently, we consider the formalism of probabilistic logic models (PLMs)~\cite{SRLBook,StaraiBook} that combine the expressive power of relational/first-order logic with the ability of probabilistic/machine learning methods to model uncertainty. Specifically, we build upon the recently successful paradigm of relational gradient-boosting for learning PLMs to address the problem of KG refinement. 

\begin{figure*}
   \centering
\includegraphics[width=.95\textwidth]{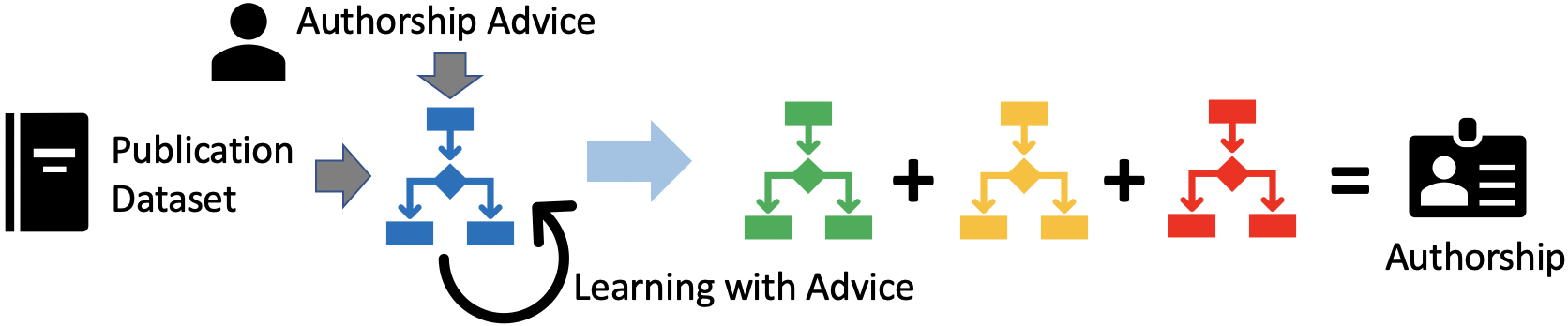}
   \caption{Overview of our approach.}
  \label{fig:teaser}
\end{figure*}

We pose the problem of KG refinement using the lens of knowledge-based learning of PLMs~\cite{OdomFrontiers}, and address it using learning from both data and knowledge. Specifically, we consider the problem of identifying scientific authorships by constructing and refining KGs.  Our approach is presented in Figure~\ref{fig:teaser}. As can be observed, we assume that there is an initial KG and some background knowledge (called {\em advice} following classic AI terminology~\cite{McCarthyAdvice,Shavlik90}) used to learn and refine the model. Our model is a set of relational regression trees~\cite{blockeel1999top,natarajan2012gradient} learned using the machinery of functional-gradient boosting. As we get more advice/data, the model can get revised accordingly. There are two important aspects of the advice: they are {\em general} where they can be applied across multiple data sets, and are provided {\em apriori} as provided before data is obtained. This ensures that the advice is independent of the specific data set and the domain expert can provide them without being biased. We demonstrate the efficacy of our learning method with advice on seven KGs.

We make the following key contributions: (1) We pose the learning and refinement of KGs as a knowledge-based probabilistic logic learning task. (2) We adapt and employ an efficient knowledge-based learning method for PLMs that effectively uses both data and knowledge to refine the KGs. (3) We show how this adaptation can be performed for the task of KG refinement. (4) We demonstrate the efficacy and efficiency of the proposed approach in seven domains.

The rest of the paper is organized as follows: after reviewing the necessary background, we explain the process of KG refinement using KBPLMs. We then present our experimental results before concluding by outlining areas of future research.

\section{Background}
We now provide the necessary background for the three different concepts that are employed in the paper -- learning weighted first-order rules using gradient-boosting, employing knowledge in the form of advice (first-order) rules and the task of author name disambiguation. 
\subsection{(Relational) Functional Gradient Boosting}
Methods that combine the expressive power of relational/first-order logic with the ability of probability theory to model uncertainty have been explored under the paradigms of Statistical Relational Learning (SRL) or Probabilistic Logic Models (PLM)~\cite{SRLBook,StaraiBook}. While powerful in representation, learning and inference remain challenging tasks for these models. In the past decade, a method based on functional-gradient boosting was proposed for these models~\cite{natarajan2012gradient,natarajan2015boosted}.

Recall that in standard conditional probability estimation (i.e., learning $P(y_i|\x_i)$), gradient descent learning calculates the gradient of the log-likelihood $\frac{\partial }{\partial \theta} \log P(y_i | \x_i; \theta)$, where $\theta$ is the model parameter. Functional gradient boosting (FGB) \cite{friedman2001greedy,dietterich2008gradient}, on the other hand, estimates the conditional probability by computing the gradient of log-likelihood w.r.t a function $\psi(\x)$ that is used to parameterize the conditional probability, typically a sigmoid function.

The sigmoid function for binary class or softmax function for multi-class that has been often adopted and used for learning PLMs  is~\cite{natarajan2015boosted,natarajan2012gradient}: $P(y_i|\x_i;\psi) = \frac{e^{ \psi(y_i,\x_i)}}{\sum_{y' }e^{ \psi(y',\x_i)}}$, where $\psi$ represents the parameters used to compute regression value $\psi(\x_i)$. Following functional gradient boosting for binary classification, the gradient of log-likelihood w.r.t $\psi(y_i=1, \x_i)$ of each training example $\x_i$ is $\Delta(y_i, \x_i) \, = \, \textit{I}(y_i=1) - \textit{P}(y_i=1|\x_i;\psi)$, where $\textit{I}$ is an indicator function which is 1 if the label of $i$-th example $y_i$ is 1, otherwise 0. The effect of $\Delta(y_i, \x_i)$ pushes predicted probability of positive examples to 1 and negative examples to 0.
As compared to standard boosting like XGBoost~\cite{chen2016xgboost}, first-order predicates are used for relational FGB instead of simple tabular representations for the features. Positive and negative examples $X=\{\x_i\}$ are ground atoms of the target predicate, for example, $\mathtt{publication(A,B)}$ and an example could be $publication("Causality", "Pearl")$. The function $\psi$ generates regression values based on the examples and related ground atoms of other predicates.

The process starts with an initial bias $\psi_0$ (usually a constant, for example, 0). At each step, a regression function is fit on the gradient $\Delta_j$. Relational regression trees (RRTs) \cite{blockeel1999top} are typically used as function representations $\psi$ for relational FGB. Hence, at the first step, $\psi_0$ is used to compute the probabilities of training examples and $\Delta_1$ can be calculated for each training example. A regression function $\hat{\Delta}_1$ is fit on the gradients of training examples. And now the parameter becomes $\psi_1 = \psi_0 + \hat{\Delta}_1$. The same operation repeats for $\psi_1$. 
Therefore, the final regression function is $\psi_k \, = \, \psi_0 + \hat{\Delta}_1 + \hat{\Delta}_2 + \dots + \hat{\Delta}_k$. As the additive aggregation of the gradient steps goes on, the regression values of positive examples increase to $+\infty$ and probability values get closer to 1. The regression values of negative examples decrease to $-\infty$ and probability values get closer to 0. After $k$ steps, $k$ trees are learned.
For relational FGB, the structure of RRTs is learned based on the purity measurement (coverage of positive examples and negative examples) for node split, which is a greedy approach with local optimum. The regression values at the leaf nodes are calculated by the gradients of training examples reaching.

\subsection{Learning with Advice}
\label{sec:advice}
Knowledge-driven learning has been pursued in standard domains from different directions \cite{kunapuli2010online,kunapuli2013guiding,towell1994knowledge}. We consider a specific form of knowledge -- {\em advice} -- that allows for a natural teacher-student guidance scenario. In standard learning settings, advice is provided over feature space to guide the learning of the models. However, in relational domains, advice is traditionally incorporated by hand-coding the structures or even parameters of the models. Many works in natural language processing, such as \cite{riedel2009markov,yoshikawa2009jointly,poon2010joint}, take such approaches, see for example the work on weak supervision \cite{chapelle2006semi}. Due to the limitation of the utilization of advice, these works are not expressive, and do not introduce potentially novel ways of interactions between advice and data. The use of advice is strictly limited to altering the data but in reality, advice is provided with the notion of altering the model. We employ advice for model refinement and not simple data reweighting. 

The model learned from source domain is used as the initial model for the target domain to boost gradients on the training data \cite{natarajan2013accelerating}. While this approach is well suited for transfer learning. The final model might drift away if advice is incorporated as the initial model and refined on the sub-optimal training data. To handle the noisy examples and consider the advice throughout the whole learning process, advice is incorporated into the gradient update of each training example in \cite{odom2015knowledge} for the model refinement.

\subsection{Author Name Disambiguation}
Traits/features of publications such as author name, venue, affiliation and etc. are useful information to identify publications from the same authors.
To decide whether two publications are from the same author or not, the pairwise similarity between publications is estimated through classification models \cite{song2015exploring,jhawar2020author,kim2019hybrid}. 
To further group all publications from the same author, the pairwise similarity distances are considered when grouping publications for each author. Plenty of clustering strategies are brought up based on the pairwise distances, such as hierarchical agglomerative clustering \cite{zhang2018name,liu2014author}, spectral clustering \cite{han2005name} and etc.

Deep neural networks (DNNs) have been adopted to extract information from publications due to their versatility. 
\cite{tran2014author} compose features manually and pass the manually-created features to DNNs. \cite{zhang2018name} firstly learn and generate representation (embeddings) of publications by DNN. A graph is then built based on the pairwise similarity distances between publications. The representation of publications is further refined by graph autoencoder. \cite{kim2019hybrid,subramanian2021s2and} estimate the pairwise similarity between publications by DNN, and train gradient boosted trees on all traits along with the pairwise similarity to improve performance.

The previous works learn embeddings and pairwise similarity of publications by DNNs, and cluster publications based on the similarity between publications. On the other hand, we utilize relational regression trees which allow for seamless integration of human knowledge by injecting relational advice. Our approach learns more explainable models, following several works on explaining additive models~\cite{craven96,natarajan2012gradient}, and we hypothesize that this allows for practical adaptation of the learning methods. 

\section{Problem Formulation}

One of our key contributions is to pose the problem of authorship as knowledge graph refinement, which is closer to practical scenario. Our problem definition is: \\
\fbox{%
    \parbox{0.98\columnwidth}{
    \textbf{\textsc{Given:}} A knowledge base $\mathcal{B}$ containing publications $\mathcal{P}$ whose authorship is known and publications $\mathcal{P}'$ whose authorship is not identified yet; \\
    \textbf{\textsc{To Do:}} Based on the existing authorship in $\mathcal{P}$, identify the authorship of other publications $\mathcal{P}'$ and complete $\mathcal{B}$.
    }%
}
\\

The most important observation that we make is that it is natural to represent such knowledge graphs using predicate logic notation. For instance, the fact that {\em Tom} co-authored a paper with {\em Amy} can be captured easily by the predicate {\em coAuthor(Tom,Amy)}. It is important to note that since there could be several authors in a paper and several papers per author, it is not efficient or effective to use a tabular representation for such rich, relational data~\cite{SRLBook,StaraiBook}. While approaches using embeddings have been used in the graph community for modeling such data for a long time, the use of a logic-based approach allows for a key property -- explainability/interpretability. Domain experts can easily understand the notations and provide appropriate inputs for such logic-based systems as evidenced by research in rule-based expert systems from a few decades ago.

Inspired by the natural representational power of logic, we transfer the knowledge base $\mathcal{B}$ to first-order using predicates.  Unless otherwise specified, from hereon, we will use upper-case letters to represent variables and lower-case letters to represent grounded items throughout the rest of the paper. A variable can refer to a set of grounded items and can be substituted by any of them. For example, $\mathtt{publication(p,a)}$ specifies that a person $\mathtt{a}$ is the author of a particular publication $\mathtt{p}$. Similarly, $\mathtt{publication(P,A)}$ states that one person $\mathtt{A}$ from the set of the authors is the author of one publication $\mathtt{P}$ from the set of publications.
We take $\mathtt{publication(P,A)}$ as our target predicate, for which we build a prediction model that is learned as a set of relational regression trees. The other observations are then converted as a set of facts (grounded predicates), over which the algorithm searches for possible weighted first-order logic rules. The predicates of facts in our authorship problem include: 
\begin{enumerate}
\item $\mathtt{author\_name(A,B)}$: author name of publication $\mathtt{A}$ is $\mathtt{B}$; \item $\mathtt{affiliation(A,B)}$: affiliation of publication $\mathtt{A}$ is $\mathtt{B}$; 
\item $\mathtt{venue(A,B)}$: publication $\mathtt{A}$ is published in venue $\mathtt{B}$; 
\item $\mathtt{reference(A,B)}$: publication $\mathtt{A}$ cites publication $\mathtt{B}$;
\item $\mathtt{coauthor(A,B)}$: coauthor of publication $\mathtt{A}$ is $\mathtt{B}$; 
\item $\mathtt{title(A,B)}$: title of publication $\mathtt{A}$ is $\mathtt{B}$.
\end{enumerate}

\section{Proposed Method}\label{subsec:method}

In this section, we present the proposed framework (Fig.~\ref{fig:teaser}) to identify the authorship, which learns a relational regression tree from knowledge graph and human advice simultaneously at each step.
Given the authorship of publications in existing knowledge graph, the learning process starts with an empty set of RRTs or an initial model of constant bias value. 
At each learning step, a new RRT will fit on the human advice, and the residual between model prediction and data. A series of RRTs will be learned following the process.
The model receives guidance from human advice while learning from the patterns presented in the data through knowledge graph refinement.
The regression value obtained from each tree will be aggregated additively to generate the probability of each example during the prediction.
We will now outline the details of the relational tree representation and how to inject human advice into the learning of the model next.

\subsection{Relational Tree Representation}\label{rrt}

\begin{figure}[htbp]
  \centering
  \includegraphics[width=0.87\linewidth]{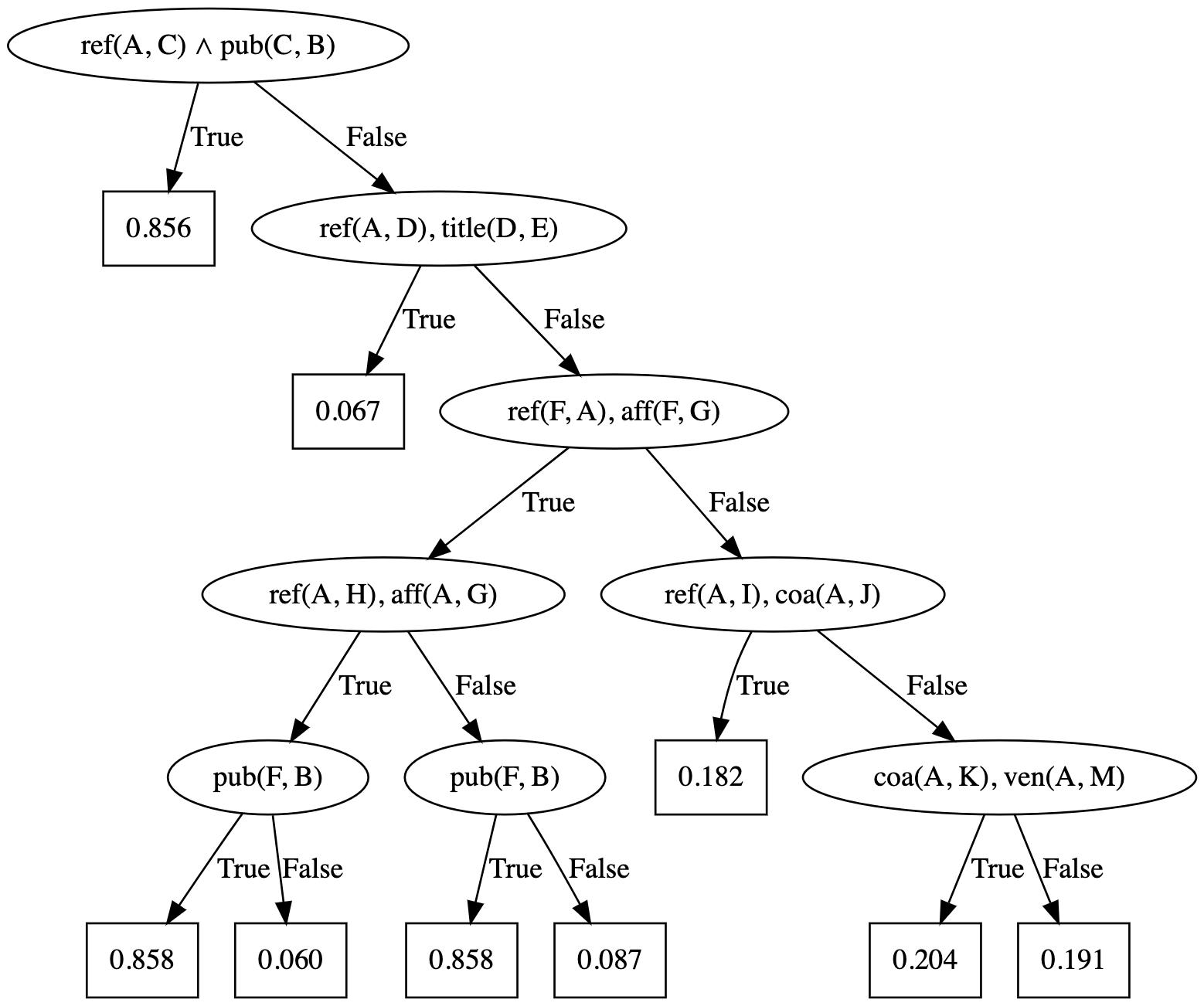}
  \caption{Tree example for publication predicate as target ($\mathtt{pub}$ - $\mathtt{publication}$, $\mathtt{ref}$ - $\mathtt{reference}$, $\mathtt{coa}$ - $\mathtt{coauthor}$, $\mathtt{aff}$ - $\mathtt{affiliation}$, $\mathtt{ven}$ - $\mathtt{venue}$, $\mathtt{aut}$ - $\mathtt{author\_name}$).}
  \label{fig:tilde-tree-example}
\end{figure}

Inspired by \cite{natarajan2012gradient}, we learn a set of RRTs to represent the relational rules, and use the learned RRTs to predict probabilities for each example of target predicate.
An example of the RRT is shown in Figure~\ref{fig:tilde-tree-example}. 
Instead of using plain features as split conditions at each node in propositional setting, in our setting,
each node of the tree consists of one or multiple first-order logic predicates, and existential quantification is applied. If there exist grounded items satisfying the node, from the sets of items, then we traverse the left branch; otherwise, we traverse the right branch. The positive/negative examples are split accordingly.
Note that each RRT can be transformed to clausal format, which are the clauses/rules that we will use in the rest of the paper. Specifically, each path in the regression tree corresponds to a conjunction of predicates and the entire set of clauses forms a decision list that will be executed in a specific order. This IF-THEN-ELSE form of representation improves the interpretability and explainability of the learned model. 
The clausal format of the RRT in Fig.~\ref{fig:tilde-tree-example} is:
\[
\small{\begin{array}{cl}
    0.856 \, \mathtt{pub(A, B)} \, \Leftarrow &\mathtt{ref(A, C)} \wedge \mathtt{pub(C, B)}. \\
    0.067 \, \mathtt{pub(A, B)} \, \Leftarrow &\mathtt{ref(A, D)} \wedge \mathtt{title(D, E)}. \\
    0.858 \, \mathtt{pub(A, B)} \, \Leftarrow &\mathtt{ref(F, A)} \wedge \mathtt{aff(F, G)} \wedge \mathtt{ref(A, H)} \wedge \mathtt{aff(A, G)} \\
    &\wedge \mathtt{pub(F, B)}. \\
    0.060 \, \mathtt{pub(A, B)} \, \Leftarrow &\mathtt{ref(F, A)} \wedge \mathtt{aff(F, G)} \wedge \mathtt{ref(A, H)} \wedge \mathtt{aff(A, G)}. \\
    0.858 \, \mathtt{pub(A, B)} \, \Leftarrow &\mathtt{ref(F, A)} \wedge \mathtt{aff(F, G)} \wedge \mathtt{pub(F, B)}. \\
    0.087 \, \mathtt{pub(A, B)} \, \Leftarrow &\mathtt{ref(F, A)} \wedge \mathtt{aff(F, G)}. \\
    0.182 \, \mathtt{pub(A, B)} \, \Leftarrow &\mathtt{ref(A, I)} \wedge \mathtt{coa(A, J)}. \\
    0.204 \, \mathtt{pub(A, B)} \, \Leftarrow &\mathtt{coa(A, K)} \wedge \mathtt{ven(A, M)}. \\
    0.191 \, \mathtt{pub(A, B)} \, \Leftarrow &.
\end{array}}
\]
Values from the leaf node of the satisfied path of each tree are used additively to calculate the probability for an example. More precisely, additive aggregation is applied for a set of trees in the model to calculate the probability.

\subsection{Advice Formulation}\label{adviceDef}

A key advantage of using a logic-based formalism is that we can simply use the same representation, i.e., Horn clauses, as advice (or prior domain knowledge) $\mathcal{A}$. Note that while the advice is also in the form of IF-THEN-ELSE rules (i.e., a decision list), the key difference is that the advice rules are not designed after obtaining the data. Instead, the advice rules are written {\em independently and before} viewing the training data. 

The obvious question is -- why not use advice as the initial model? As prior results in the literature \cite{OdomFrontiers} show, the use of advice as the first-tree in boosting can result in data overwhelming the advice and this is particularly problematic. Instead, following standard successes in machine learning, we use advice as a form of {\em regularization} to the model and allows for explicitly handling noisy data. 

We now define advice constraint and advice set based on the definitions of Odom et al.~\cite{OdomFrontiers}. 

\begin{definition}[Advice Constraint $\mathcal{R}$]
    Denote by bold $\x_i$ the $i$-th example. An advice constraint is evaluated on each example $\x_i$, which is formulated by a Horn clause: $l(\x_i) \Leftarrow {\land}_{j} \; predicate_j (\x_i )$. That is, if there exist facts in KG that satisfy all predicates, then the corresponding label will be yielded.
\end{definition}
\begin{definition}[Advice Set $\mathcal{A}$]
    An advice set is a set of advice rules, each of which is composed of an advice constraint $\mathcal{R}$ and label preferences $l\pm$: $\langle \mathcal{R}$, $l+$, $l- \rangle$. When the advice constraint $\mathcal{R}$ is satisfied by an example, the preferred label $l+$ is assigned to the example; otherwise, the avoided label $l-$ is assigned.
\end{definition}

\label{advice}
Concretely, we use the following advice for the authorship prediction. Two specific aspects of these rules need to be highlighted: (1) The rules are {\em general} in that they are not specific to any data set and instead used across multiple data sets. (2) The rules are provided {\em apriori} in that the data is not used to generate these rules. Instead, the rules are designed by us as domain experts in the area. We argue that this type of knowledge injection is much more natural when compared to weak supervision \cite{chapelle2006semi} or distant supervision \cite{mintz2009distant}.

\begin{enumerate}
    \item Two papers are from the same author if their author names are the same:
    \[
    \small{\begin{array}{cl}
    \mathtt{pub(A, B)} \, \Leftarrow &\mathtt{aut(A, Q)} \wedge \mathtt{aut(R, Q)} \wedge \mathtt{pub(R, B)}. 
    \end{array}}
    \]
    \item Two papers are from the same author if their affiliations are the same:
    \[
    \small{\begin{array}{cl}
    \mathtt{pub(A, B)} \, \Leftarrow \mathtt{aff(A, E)} \wedge \mathtt{aff(F, E)} \wedge \mathtt{pub(F, B)}. 
    \end{array}}
    \]
    \item Two papers are from the same author if they are published in the same venue:
    \[
    \small{\begin{array}{cl}
    \mathtt{pub(A, B)} \, \Leftarrow \mathtt{ven(A, C)} \wedge \mathtt{ven(D, C)} \wedge \mathtt{pub(D, B)}. 
    \end{array}}
    \]
    \item Two papers are from the same author if there is some reference relation between them:
    \[
    \small{\begin{array}{cl}
    \mathtt{pub(A, B)} \, \Leftarrow &\mathtt{ref(A, G)} \wedge \mathtt{pub(G, B)}. \\
    \mathtt{pub(A, B)} \, \Leftarrow &\mathtt{ref(H, A)} \wedge \mathtt{pub(H, B)}. \\
    \mathtt{pub(A, B)} \, \Leftarrow &\mathtt{ref(A, I)} \wedge \mathtt{ref(J, I)} \wedge \mathtt{pub(J, B)}. \\
    \mathtt{pub(A, B)} \, \Leftarrow &\mathtt{ref(K, A)} \wedge \mathtt{ref(L, K)} \wedge \mathtt{pub(L, B)}. \\
    \mathtt{pub(A, B)} \, \Leftarrow &\mathtt{ref(A, M)} \wedge \mathtt{ref(M, N)} \wedge \mathtt{pub(N, B)}. 
    \end{array}}
    \]
\end{enumerate}
Each piece of advice is general enough and solely cannot identify the authorship precisely, but provide valuable information to guide the learning of the model. We now discuss how to learn the model on top of these advice rules. 
\subsection{Injecting Advice}\label{inject-advice}
As mentioned earlier in section \ref{sec:advice}, one simple way to incorporate advice is to transfer the rules in advice to the initial model (first tree of the boosted-tree model), and then use data to further generate subsequent models (series of trees) to refine the decision boundary. However, the main drawback of this method is that the further learning of subsequent models by fitting on the data may eliminate the influence of the advice, especially under the noisy data case. To incorporate advice at every learning step, we consider including advice as part of the objective function (similar to regularization) into the gradient update of the model parameters as \cite{odom2015knowledge}.
We inject advice into the gradient of each example, which affects not only the estimation of the values of parameters but the learning of the model structure.

Denote by $y_i$ the label of the $i$-th instance of authorship, whose value is true or false. Inspired by \cite{odom2015knowledge}, we incorporate advice as cost into the log-likelihood objective function, and therefore the advice is included in the gradient update of the model parameters. We have the following gradient calculation, which incorporates human advice by counting the satisfied advice of each example:
\begin{align*}
\Delta(y_i, \x_i) \, = \, \underbrace{\textit{I}(y_i=1) - \textit{P}(y_i=1|\x_i;\psi)}_{\textrm{Gradient from Data}}\\ 
+ \underbrace{\lambda \cdot [ n_t (\x_i) - n_f (\x_i) ]}_{\textrm{Gradient from Advice}},
\end{align*}
where $\lambda$ is a scalar value, used to trade off between data and advice. Since we focus on whether the label value of each example $\x_i$ is true or false, we adopt two types of advice in this paper: 1) positive advice prefers the label value of {\em true authorships} to be true; 2) negative advice prefers the label value of {\em non-existent authorships} to be false. $n_t$ and $n_f$ count the number of positive and negative advice satisfied by the example, respectively. The difference between positive advice counting and negative advice counting decides the gradient from advice.

Since the scaling is a better way to sort of normalize the gradient to consider the trade-off between data and advice, we rewrite the gradient through scaling the gradient by $\eta = \alpha$ and set $\lambda = (1-\alpha)/\eta$. Hence, we can get
\begin{align*}
\eta \cdot \Delta(y_i, \x_i) \, = \, &\alpha \cdot [\textit{I}(y_i=1) - \textit{P}(y_i=1|\x_i;\psi)] \\
&+ (1 - \alpha) \cdot [ n_t (\x_i) - n_f (\x_i) ]
\end{align*}

During implementation, we simplify the advice weight by assigning 1.0 gradient to examples satisfying any given advice, and 0.0 gradient to the rest of the examples:
\begin{align} \label{eq:gradient-advice}
    \Delta'(y_i, \x_i) \, = \, &\alpha \cdot [\textit{I}(y_i=1) - \textit{P}(y_i=1|\x_i;\psi)] \nonumber\\
&+ (1 - \alpha) \cdot \textit{I}(\exists \mathcal{R} \in \mathcal{A}, \; \mathcal{R}(\x_i))
\end{align}
Due to the fact that assigning the same advice weight to each piece of advice might be inappropriate, and the exact weight for each piece of advice is hard to decide individually, we use advice as sign signal to help adjust the decision boundary. This allows us to not trade off the advice against each other but to use them as general guidance when learning the trees. The intuition is that these are small reward signals (whose weight is governed by $\alpha$) that reinforce the choices made by the boosted model. It should be mentioned that we do not explicitly provide negative advice for this domain. While in medical domains such advice could be intuitive to provide (for instance absence of a risk factor might point to the absence of the diseases), it is not straightforward to provide such negative advice in this domain and hence we avoid them.

\section{Experiments}
We now present the performance of our proposed method in different aspects on various datasets.

\subsection{Experimental Setup}

\noindent{\textbf{Datasets:}}\;
We conduct experiments on seven scientific publication datasets: AMiner \cite{zhang2018name}, ArnetMiner \cite{tang2011unified}, INSPIRE \cite{louppe2016ethnicity}, KISTI \cite{kang2011construction}, PubMed \cite{song2015exploring}, QIAN \cite{qian2015dynamic}, zbMATH \cite{muller2017data}. The numbers of publications and authors are shown in Table~\ref{tab:datasets}.
\begin{table}[htbp]
  \caption{Statistics of Datasets}
  \label{tab:datasets}
  \begin{center}
  \begin{tabular}{crr}
    \toprule
    Dataset & \#Publications & \#Authors\\
    \midrule
    AMiner & 157,448 & 20,609\\
    ArnetMiner & 7,144 & 796\\
    INSPIRE & 536,564 & 57,978\\
    KISTI & 40,383 & 7,161\\
    PubMed & 2,871 & 719\\
    QIAN & 6,542 & 1,321\\
    zbMATH & 15,181 & 4,025\\
  \bottomrule
\end{tabular}
\end{center}
\end{table}

\begin{figure}[htbp]
\begin{subfigure}[b]{0.49\linewidth}
    \includegraphics[width=\linewidth]{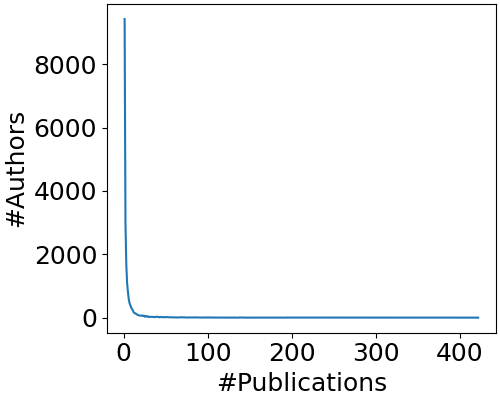}
    \caption{ }
    \label{fig:aminer-author-publication-a}
\end{subfigure}%
\hspace{\fill}
\begin{subfigure}[b]{0.49\linewidth}
    \includegraphics[width=\linewidth]{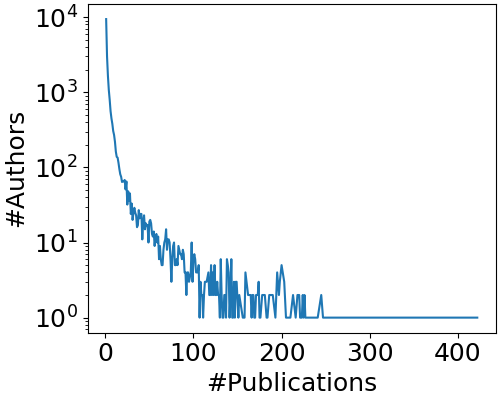}
    \caption{ }
    \label{fig:aminer-author-publication-b}
\end{subfigure}
\caption{Distribution of authors with respect to publication numbers on AMiner dataset.}
\label{fig:aminer-author-publication}
\end{figure}

To explore the characteristics of the datasets, we plot the distribution of authors with respect to publication amounts on AMiner dataset, and the results are shown in Figure~\ref{fig:aminer-author-publication-a}, \ref{fig:aminer-author-publication-b} (take the logarithm of author amounts). It is observed that {\em authors' publications follow a power-law distribution: few authors publish the most of papers and several authors only publish very few papers}. 

\noindent{\textbf{Datasets Preprocessing:}} We perform the following steps:
\begin{enumerate}
    \item Remove publications whose author\_id is None, which means the author\_id is missing.
    \item Remove publications associated with multiple first authors.
    \item Remove author\_id who published fewer than 10 papers.
    \item In case there are still too many authors and publications after steps (1)-(3), we sample 10\% of the authors and their publications from the dataset.
    \item Split the data into training and testing data with a ratio of 80\%:20\%.
    \item Format strings of author names, affiliations, venues and etc. in order to prepare predicates for boosted relational tree learner.
    \item A positive example is that a paper\_id is published by an author\_id while a negative example is not published by. When generating negative examples/pairs from the positive examples, we randomly pick negative examples such that the number of negative examples is twice the number of positive examples. 
\end{enumerate}
As an example, for AMiner dataset, after removing authors whose publications are less than 10, there are only 2,809 (13.63\%) authors and 108,955 (69.20\%) papers left, which reflects the same tendency we observe in Figure \ref{fig:aminer-author-publication}.

Note that domain knowledge (either as regularization or as domain constraints) is most useful when the data is noisy or mislabeled or completely missing. To simulate these scenarios, we perform the following:

\begin{itemize}
    \item \textbf{Random Noisy Labels:} To generate the random noisy labels, we randomly flipped 50\% of positive examples and 25\% of negative examples to keep the ratio of negative and positive examples roughly 2:1.
    \item \textbf{Random Missing Labels:} To generate the missing labels, we randomly sample 50\% of positive examples and 25\% of negative examples, and regard the labels of those examples as unobserved.
\end{itemize}

\begin{figure*}[t]
  \centering
  \includegraphics[width=0.72\textwidth]{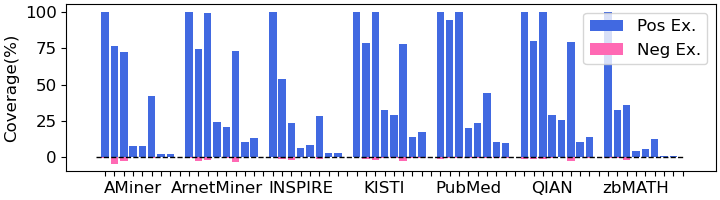}
  \caption{Coverage of each piece of advice (8 pieces) on each dataset for positive and negative Examples.}
  \label{fig:advice-coverage}
\end{figure*}

\noindent{\textbf{Questions:}}\;
In our experiments, we aim to answer the following questions empirically:
\begin{itemize}
    \item[\textbf{Q1:}] Is the advice based framework effective for author disambiguation task across all datasets?
    
    \item[\textbf{Q2:}] Is the algorithm robust to the choice of the advice weight parameter $\alpha$ in the disambiguation task?
    \item[\textbf{Q3:}] Does our framework perform comparably to stronger baselines?
\end{itemize}

We will resolve these questions in the following parts. Specifically, we verify the quality of each piece of advice (\textbf{Q1}), evaluate the importance of the advice weight (\textbf{Q2}), and examine the learned model with and without advice as against the baselines (\textbf{Q3}). In addition to training the model on each dataset separately, we also perform a generalization experiment by training the model on one subset of the seven datasets and testing it on another subset.
In experiments, we use evaluation metrics AUC ROC (receiver operating characteristic curve) and AUC PR (precision recall curve) to quantify the model performance, where AUC is the area under the curve.

\subsection{Measure Advice Quality}

One way of measuring the quality of advice is to compute the example coverage on the training examples of each dataset. We calculate the example coverage for each piece of advice individually. Ideally, we expect that the advice can be applied to a larger number of positive examples and a smaller number of negative examples. If the advice covers comparably an equal number of positive and negative examples, then the advice is not particularly useful for learning a discriminative classifier. On the other hand, if the advice covers only positive examples and no negative examples, then the advice might be perfect and  in the case of clean data, the same information can already be obtained from the data. However, with noise, this advice can be quite useful. Figure~\ref{fig:advice-coverage} presents the positive example coverage (blue bars) and negative example (pink bars) coverage (satisfied) of advice in the same order as listed in section 4.2. The analysis of Figure~\ref{fig:advice-coverage} reveals that certain advice pieces exhibit a remarkably high coverage percentage of positive examples. Moreover, we observe that our advice can achieve significantly higher coverage for positive examples compared to negative examples. By incorporating multiple pieces of advice, the model is able to learn a more precise decision boundary within the example space, which cannot be learned solely from the data. 

\begin{figure}[ht]
  \centering
  \includegraphics[width=0.75\linewidth]{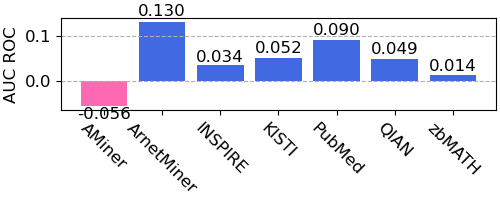}
  \caption{AUC ROC under noisy label case.}
  \label{fig:single-noisy-roc}
\end{figure}

\begin{figure}[ht]
  \centering
  \includegraphics[width=0.75\linewidth]{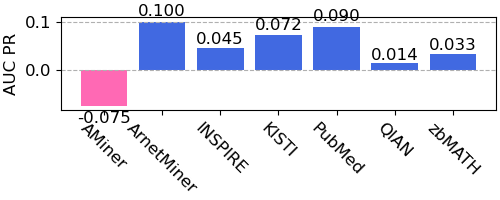}
  \caption{AUC PR under noisy label case.}
  \label{fig:single-noisy-pr}
\end{figure}

\begin{figure}[ht]
  \centering
  \includegraphics[width=0.75\linewidth]{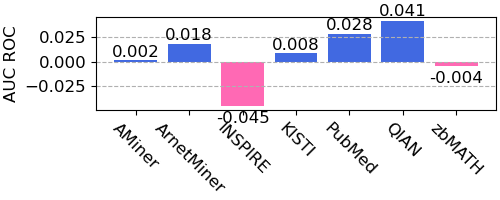}
  \caption{AUC ROC under missing label case.}
  \label{fig:single-missing-roc}
\end{figure}

\begin{figure}[ht]
  \centering
  \includegraphics[width=0.75\linewidth]{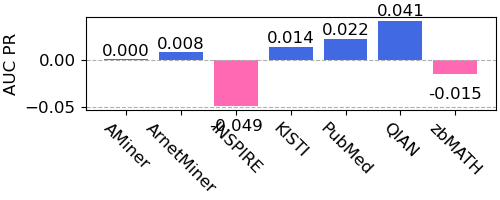}
  \caption{AUC PR under missing label case.}
  \label{fig:single-missing-pr}
\end{figure}

To explore the performance of the set of advice on each individual dataset with noisy or missing labels, we set the advice weight as 0.5. Figures~\ref{fig:single-noisy-roc} and \ref{fig:single-noisy-pr} illustrate the results obtained for noisy labels case, while Figures~\ref{fig:single-missing-roc} and \ref{fig:single-missing-pr} display the results for the case of missing labels. 
The bars in these figures show the difference between the performance of models learned with and without advice. 
From these results, we observe that in general, the incorporation of advice enhances the performance of the learned model with no tuning on advice weight. However, when using an advice weight of 0.5 without additional tuning, the performance may be adversely affected on some datasets.
However, validation of advice weight can help achieve the optimal performance which can be observed from the results in the following part.
Based on these findings, we can conclude that the given advice is sufficiently effective for our task of author disambiguation. Its ability to provide higher coverage of positive examples and performance improvement on individual datasets demonstrate the effectiveness of the advice in learning a more effective model, thus answering \textbf{Q1} affirmatively.



\subsection{Robustness to the Weight Parameter}\label{sec:robust}

\subsubsection{Single Datasets}

\begin{figure}[h]
  \centering
  \includegraphics[width=0.75\linewidth]{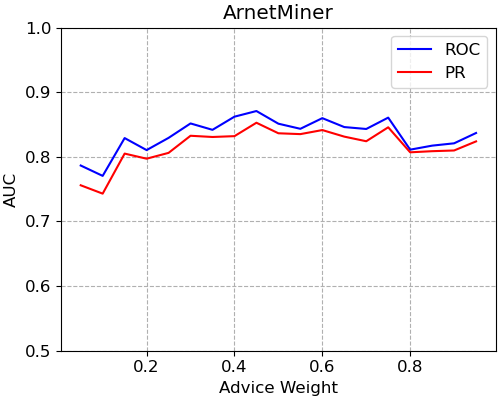}
  \caption{Model performance on ArnetMiner with varied advice weight.}
  \label{fig:advice-arnetminer}
\end{figure}

\begin{figure}[h]
  \centering
  \includegraphics[width=0.75\linewidth]{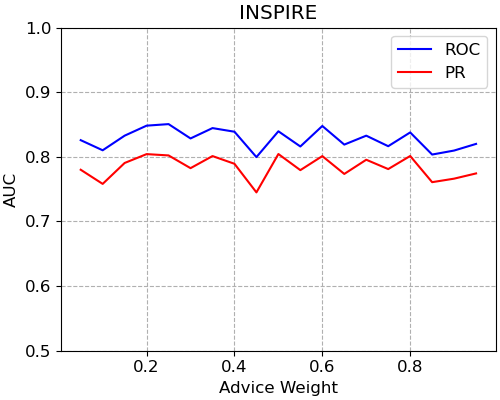}
  \caption{Model performance on INSPIRE with varied advice weight.}
  \label{fig:advice-pubmed}
\end{figure}

To further evaluate the influence of advice weight, we conduct experiments on each single dataset with missing label and advice weights varied in [0.05, 0.95]. The results for ArnetMiner and INSPIRE datasets are presented in Figures~\ref{fig:advice-arnetminer} and \ref{fig:advice-pubmed}, respectively. 
Although the performance fluctuates in the range [0.05, 0.95], when the advice weight is within the range [$0.25$, $0.75$], there is only a small difference in the performance of models learned with advice, which indicates that the model is robust within this advice weight range. To achieve optimal performance, we may need to validate and fine-tune the advice weight. 


\subsubsection{Mixed Datasets}

\begin{figure}[h]
  \centering
  \includegraphics[width=0.75\linewidth]{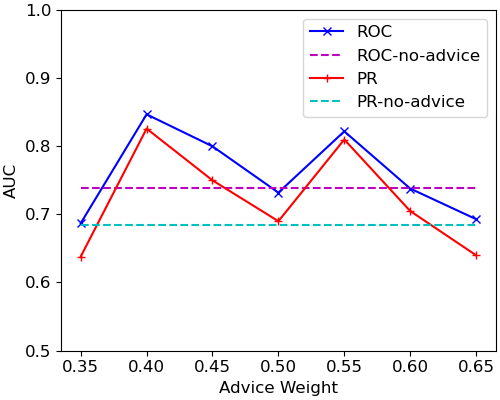}
  \caption{Mixed data: train on 3 and test on 4.}
  \label{fig:train3test4}
\end{figure}

\begin{figure}[h]
  \centering
  \includegraphics[width=0.75\linewidth]{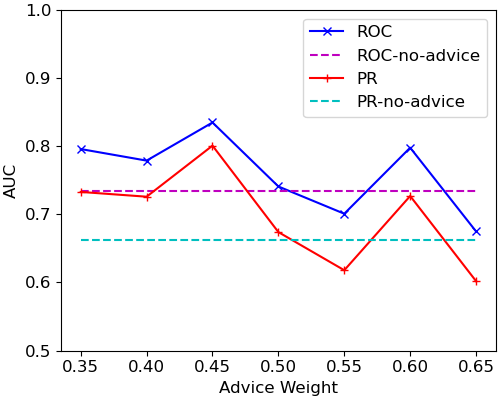}
  \caption{Mixed data: train on 7 and test on 7.}
  \label{fig:train7test7}
\end{figure}

To further assess the impact of advice weight on mixed datasets, we employ the training data merged from three datasets (AMiner, ArnetMiner, INSPIRE) for model training and conduct testing using testing data from the remaining four datasets. It is noted that all the data used in this analysis are with noisy labels, and the advice weight is varied to observe the fluctuation in performance. The corresponding results on the two evaluation metrics AUC ROC and AUC PR are presented in Figure~\ref{fig:train3test4} for the models learned with advice (solid curves) and without advice (dashed horizontal lines). Additionally, we also train and test the model using the training and testing data of all seven datasets, and the results are displayed in Figure~\ref{fig:train7test7}. 
The analysis of these results indicates that the model trained with advice outperforms the one trained without advice to a significant extent, except for some specific advice weight choices. 
These observations demonstrate that the advice weight can have a reasonable impact on the performance of the learned model, and finding the right balance is important to ensuring the effectiveness of incorporating the advice.
However, overall, having a reasonable advice weight yields within a certain threshold of the optimal performance demonstrating the robustness of the model. (\textbf{Q2})


\subsection{Generalization Performance}

In this group of experiments, we perform training on the data merged from four datasets (INSPIRE, KISTI, PubMed and zbMATH), testing on two datasets (AMiner and QIAN), and validating on ArnetMiner dataset. 

\subsubsection{Performance Analysis.}

As we observed in section \ref{sec:robust}, due to the trade-off between advice and data, the reasonable advice weight range is from $0.25$ to $0.75$. A larger advice weight makes the model lean more towards the advice, while a smaller advice weight pushes the model to rely more on the data. To find the optimal advice weight, an exploration was carried out over this advice weight range, utilizing an interval of 0.05. The results indicate that to achieve the highest AUC on validation data, an advice weight of 0.75 is the most effective for the noisy label case, while an advice weight of 0.60 performs best for the missing label case. Subsequently, the model's performance is evaluated on the testing data using the corresponding best advice weights, and the results are presented in Table~\ref{tab:mixed-advice-weight}. In the context of our model, the denotation "NA" corresponds to the case where no advice is incorporated, while "AW" represents our model trained with advice using the advice weight for noisy (0.75) and missing label (0.60) cases, respectively. We observe that the performance of the learned model improves with the help of the human advice under both the noisy label case and missing label case, compared to the model learned solely from the data. Furthermore, the disparity in performance achieved between learning with and without advice is more pronounced for the noisy label case.

\textbf{Comparisons with strong baselines.} To answer \textbf{Q3}, we compare the performance of our proposed method to a neural network baseline, DNN, proposed by previous works for author name disambiguation. DNN is designed to learn embeddings based on the facts/features of publications, which is less susceptible to the influence of authorship information compared to our approach. However, human knowledge is generally not easy to be incorporated into DNN. Much more efforts are needed to carefully craft human knowledge into the structure and parameters of DNN. While, for our framework, advice can be included just by simple rules as in section~\ref{adviceDef}. The results are summarized in Table~\ref{tab:mixed-advice-weight}. Although the authorship information varies for the noisy and missing label cases, the facts/features about the publications are the same, resulting in the performance of DNN under these two cases being quite close. Besides, our method with advice outperforms DNN under both noisy and missing label cases. The reason is that DNN generally requires a very large amount of data to achieve good performance. Under the noisy label case, if no advice is provided, our learned model of RRTs may suffer the effect of the noise in authorship information, and gets slightly lower performance than DNN. This discrepancy can be attributed to our method's reliance on accurate authorship information in the training data. Nonetheless, the inclusion of advice significantly improves the performance of our model in such situation. Besides, we also evaluated another baseline model, called MLN-B \cite{khot2015gradient}, in our comparisons. Instead of using tree representation at each boosting step as our model, MLN-B learns a set of plain clauses at each step. It can be observed that our tree representation consistently outperforms the plain clauses representation of MLN-B. These observations show the superiority of our proposed method.

\begin{table}
  \centering
  \caption{Performance on Mixed Testing Datasets}
  \label{tab:mixed-advice-weight}
  \begin{tabular}{c|c|cc|cc}
    \toprule
    \multirow{2}{*}{Model}&\multirow{2}{*}{Setting}&\multicolumn{2}{c|}{Noisy}&\multicolumn{2}{c}{Missing}\\
    \cline{3-6}
    &&ROC&PR&ROC&PR\\
    \hline
    DNN&-&0.657&0.631&0.657&0.623\\
    MLN-B&-&0.594&0.521&0.695&0.689\\
    \hline
    \multirow{4}{*}{Our Model}&NA&0.614&0.539&0.818&0.808\\
    &AW&\textbf{0.724}&\textbf{0.637}&\textbf{0.848}&\textbf{0.831}\\  
    \cline{2-6}
     & \textbf{Improvement} & \multirow{2}{*}{\textbf{17.9\%}} & \multirow{2}{*}{\textbf{18.2\%}} & \multirow{2}{*}{\textbf{3.7\%}} & \multirow{2}{*}{\textbf{2.8\%}} \\ 
     & \textbf{of AW to NA}&&&&\\
  \bottomrule
\end{tabular}
\end{table}

\subsubsection{Size of Final Tree} 
\begin{figure}[htbp]
  \centering
  \includegraphics[width=0.75\linewidth]{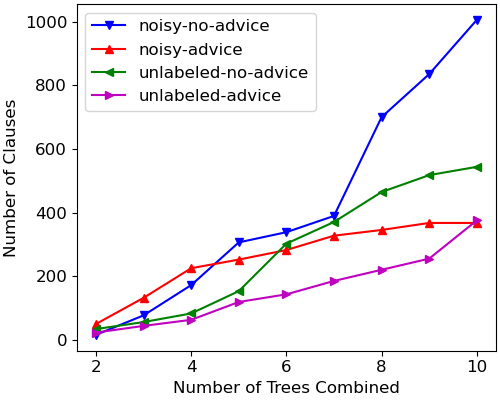}
  \caption{Number of clauses in combined Tree through CoTE.}
  \label{fig:cote-num}
\end{figure}

\begin{figure}[h]
  \centering
  \includegraphics[width=0.75\linewidth]{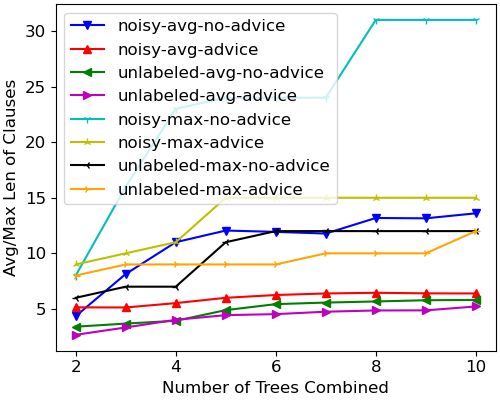}
  \caption{Avg/Max length of clauses in combined tree of CoTE.}
  \label{fig:cote-avg-max}
\end{figure}

The CoTE algorithm \cite{yan2022explainable} is utilized to combine the learned boosted-tree models under four scenarios -- noisy label and missing label cases with and without advice, and produce one combined final tree for each scenario. The statistics are shown in Figure~\ref{fig:cote-num} and \ref{fig:cote-avg-max}. 
Figure~\ref{fig:cote-num} draws the number of clauses in the final combined tree. It is observed that for both noisy and missing label cases, the inclusion of advice leads to a significant reduction in the number of clauses in the combined tree. Notably, the impact is more pronounced in the noisy label case, where the number of clauses learned with advice is less than one-third of that observed in the model without advice. Therefore, the advice can effectively help reduce the number of clauses caused by noise in the data, and elicit guidance from the unlabeled examples.

Fig.~\ref{fig:cote-avg-max} shows the average and maximum length of clauses in the final combined tree.
By analyzing the average length of clauses in the four scenarios, we observe that the final combined tree learned with advice contains shorter clauses for both the noisy and missing label cases, and this discrepancy is particularly notable for the noisy label case. 
On the other hand, the maximum length of clauses of the final tree learned without advice under noisy label case is extremely large than the other three scenarios. 
This may be due to the impact of the noise in the data. In general, reasonably smaller (max/avg) lengths of clauses are preferred. Besides, under the missing label case, the maximum length of clauses is the same for the final combined tree learned with and without advice. 

\subsubsection{Top Clauses/Rules of Final Tree}
After combining the boosted-tree models by the CoTE algorithm, we keep the top three clauses in the final combined tree for each scenario, after removing empty clauses or clauses with much more negative example coverage than positive example coverage.

\begin{enumerate}
    \item \textbf{Noisy Label Case:} We compare the final trees of the models learned with advice ({\it Advice}) and without advice ({\it No-advice}).
    \begin{enumerate}
        \item {\it No-advice}
            \[
    \small{\begin{array}{cl}
    \mathtt{pub(A, B)} \, \Leftarrow &\mathtt{ref(A, C)} \wedge \mathtt{coa(A, D)}. \\
    \mathtt{pub(A, B)} \, \Leftarrow &\mathtt{ref(A, C)} \wedge \mathtt{coa(A, D)} \wedge \mathtt{ref(A, E)} \\
    &\wedge \mathtt{coa(E, F)}. \\
    \mathtt{pub(A, B)} \, \Leftarrow &\mathtt{aff(A, C)} \wedge \mathtt{coa(A, D)} \wedge \mathtt{pub(J, B)}.
    \end{array}}
            \]
        \item {\it Advice}
            \[
    \small{\begin{array}{cl}
    \mathtt{pub(A, B)} \, \Leftarrow &\mathtt{aff(C, D)} \wedge \mathtt{pub(C, B)} \wedge \mathtt{aff(A, D)}. \\
    \mathtt{pub(A, B)} \, \Leftarrow &\mathtt{ven(A, C)} \wedge \mathtt{ven(D, C)} \wedge \mathtt{pub(D, B)}. \\
    \mathtt{pub(A, B)} \, \Leftarrow &\mathtt{coa(A, C)} \wedge \mathtt{coa(D, C)} \wedge \mathtt{pub(D, B)}. 
    \end{array}}
            \]
    \end{enumerate}

    We observe that the model without advice struggles to learn meaningful rules due to the significant noise in the training data. However, when advice is provided, the model can adopt the rules in advice and learn more sensible rules.
    Except for the clauses in advice, the model with advice is able to discover new clauses from the noisy data that two publications are from the same author if they have the same name of coauthors.

    \item \textbf{Missing Label Case:} Similarly, the final trees of the models learned with advice ({\it Advice}) and without advice ({\it No-advice}) are compared.
    \begin{enumerate}
        \item {\it No-advice}
            \[
    \small{\begin{array}{cl}
    \mathtt{pub(A, B)} \, \Leftarrow &\mathtt{coa(C, D)} \wedge \mathtt{pub(C, B)} \wedge \mathtt{coa(A, D)}. \\
    \mathtt{pub(A, B)} \, \Leftarrow &\mathtt{ref(A, C)} \wedge \mathtt{ven(A, D)} \wedge \mathtt{aff(A, E)}  \\
    &\wedge \mathtt{coa(A, F)}
    \wedge \mathtt{coa(G, F)} \wedge \mathtt{pub(G, B)}. \\
    \mathtt{pub(A, B)} \, \Leftarrow &\mathtt{ven(A, C)} \wedge \mathtt{coa(D, E)} \wedge \mathtt{pub(D, B)} \\
    &\wedge \mathtt{coa(A, E)} 
    \wedge \mathtt{aff(A, F)}. 
    \end{array}}
            \]
        \item {\it Advice}
            \[
    \small{\begin{array}{cl}
    \mathtt{pub(A, B)} \, \Leftarrow &\mathtt{ven(A, C)} \wedge \mathtt{pub(D, B)} \wedge \mathtt{aff(A, E)} \\ 
    & \wedge \mathtt{aff(D, E)} 
    \wedge \mathtt{coa(A, F)} \wedge \mathtt{pub(G, B)} \\
    &\wedge \mathtt{coa(G, F)} \wedge \mathtt{ref(A, H)}. \\
    \mathtt{pub(A, B)} \, \Leftarrow &\mathtt{ven(A, C)} \wedge \mathtt{pub(D, B)} \wedge \mathtt{aff(A, E)} \\
    &\wedge \mathtt{aff(D, E)} 
    \wedge \mathtt{coa(A, F)} \wedge \mathtt{pub(G, B)} \\
    &\wedge \mathtt{coa(G, F)}. \\
    \mathtt{pub(A, B)} \, \Leftarrow &\mathtt{aff(A, C)} \wedge \mathtt{pub(D, B)} \wedge \mathtt{aff(D, C)} \\
    &\wedge \mathtt{coa(A, E)}. 
    \end{array}}
            \]
    \end{enumerate}

    We can observe that the model with advice injected learns better combinations of simple rules. For example, in order to decide if paper $\mathtt{A}$ is published by author $\mathtt{B}$, the first rule of {\it Advice} model requires that 1) there exists a publication of author $\mathtt{B}$, which has the same author affiliation as paper $\mathtt{A}$, and 2) a publication of author $\mathtt{B}$ has the same coauthor name as publication $\mathtt{A}$.
    These rules can recognize the positive examples more precisely against the negative examples.
\end{enumerate}

\section{Conclusion}

We consider the problem of author disambiguation and knowledge graph refinement, and learn relational regression trees using functional gradient boosting. We use human knowledge as advice in the form of first-order clauses to refine the relational regression trees through the gradient update. 
The learned model benefits from the relational advice under both the noisy label and missing label cases. By combining the trees into one tree, we examine semantically the quality of the top rules brought by human advice.
Extending this work to different knowledge graphs is an interesting future direction. Understanding the use of other domain constraints such as fairness and safety constraints in building knowledge graphs for sensitive data is a necessary next step. Finally building a model that allows humans to interactively provide advice while understanding the model's details remains an exciting future direction.

\bibliographystyle{IEEEtran}
\bibliography{references}

\end{document}